\documentclass[12pt,a4paper]{article}

\usepackage[english]{babel}

\usepackage{graphicx}
\usepackage{amsmath,amsfonts,amssymb}
\usepackage{authblk}
\usepackage{amsbsy}
\usepackage{mathrsfs} 
\usepackage{varioref,hyperref}
\usepackage[shortlabels]{enumitem}
\usepackage{bm}
\usepackage{float}
\usepackage{algorithm}
\usepackage{algorithmicx}
\usepackage{algpseudocode}
\usepackage{listings}
\lstdefinestyle{cpp}{
	language=c++,
	breaklines=true,
	basicstyle={\small\ttfamily},
	showstringspaces=false,
	showtabs=false,
	keepspaces=false,
	columns=fullflexible,
	tabsize=1
}
\usepackage{placeins}
\usepackage[dvipsnames]{xcolor}

\usepackage[font=small]{caption}
\usepackage{epsfig}
\usepackage{array}
\usepackage{verbatim}
\usepackage{graphicx}
\usepackage{color}
\usepackage{tabularx}
\usepackage{bm}
\usepackage{booktabs}
\usepackage{subfig}
\usepackage{stmaryrd}
\usepackage{textcomp}

\usepackage{array}

\usepackage{multirow}
\usepackage{tabularx}
\usepackage{url}

\begin{document}

\title{A Supervised Machine-Learning Approach For Turboshaft Engine Dynamic Modeling Under Real Flight Conditions}

\markboth{Paniccia et al.}{A Supervised Machine-Learning Approach For Turboshaft Engine  \ldots}

\author[1]{Damiano Paniccia\thanks{{ Corresponding author: damiano.paniccia@leonardo.com}}}
\author[1]{Francesco Aldo Tucci}
\author[1]{Joel Guerrero}
\affil[1]{\small Leonardo Labs, Leonardo S.p.A., via R. Pieragostini 80, 16151 Genoa, Italy}
\author[2]{Luigi Capone}
\affil[2]{\small Leonardo S.p.A., via R. Pieragostini 80, 16151 Genoa, Italy}
\author[3]{Nicoletta Sanguini}
\author[3]{Tommaso Benacchio\thanks{{ Now at: Weather Research, Danish Meteorological Institute, Sankt Kjelds Plads 11, 2100 Copenhagen Denmark}}}
\author[3]{Luigi Bottasso}
\affil[3]{\small Leonardo Labs, Leonardo S.p.A., via G. Agusta 520, 21017 Samarate, Italy}
\date{}

\maketitle

\newpage
\vfill
{\noindent\footnotesize This work has not yet been peer-reviewed and is provided by the contributing authors as a means to ensure timely dissemination of scholarly and technical work on a noncommercial basis. Copyright and all rights therein are maintained by the authors or by other copyright owners. It is understood that all persons copying this information will adhere to the terms and constraints invoked by each author's copyright. This work may not be reposted without explicit permission of the copyright owner. This work has been submitted for publication. Copyright in this work may be transferred without further notice.}

\pagebreak

\abstract{Rotorcraft engines are highly complex, nonlinear thermodynamic systems that operate under varying environmental and flight conditions. Simulating their dynamics is crucial for design, fault diagnostics, and deterioration control phases, and requires robust and reliable control systems to estimate engine performance throughout flight envelope. 
Numerical simulations allow for an accurate assessment of engines behaviors in both steady and unsteady scenarios by means of physics-based and in-depth mathematical descriptions. However, the development of such detailed physical models is a very challenging task due to the complex and entangled physics driving the engine.
In this scenario, data-driven machine-learning techniques are of great interest to the aircraft engine community, due to their ability to describe nonlinear systems' dynamic behavior and enable online performance estimation, achieving excellent results with accuracy competitive with the state of the art.
In this work, we explore different Neural Network architectures to model the turboshaft engine of Leonardo’s AW189P4 prototype, aiming to predict the engine torque. The models are trained on an extensive database of real flight tests. This dataset involves a variety of operational maneuvers performed under different flight conditions, providing a comprehensive representation of the engine's performance. To complement the neural network approach, we apply Sparse Identification of Nonlinear Dynamics (SINDy) to derive a low-dimensional dynamical model from the available data, describing the relationship between fuel flow and engine torque. The resulting model showcases SINDy's capability to recover the actual physics underlying the engine dynamics and demonstrates its potential for investigating more complex aspects of the engine using the SINDy approach. 
This paper details development steps and prediction results of each model, proving that data-driven engine models can exploit a wider range of parameters than standard transfer function-based approaches, enabling the use of trained schemes to simulate nonlinear effects in different engines and helicopters.}




\newpage

\section{Introduction}
\begin{figure}[h!]
\centering
\includegraphics[width=0.7\textwidth]{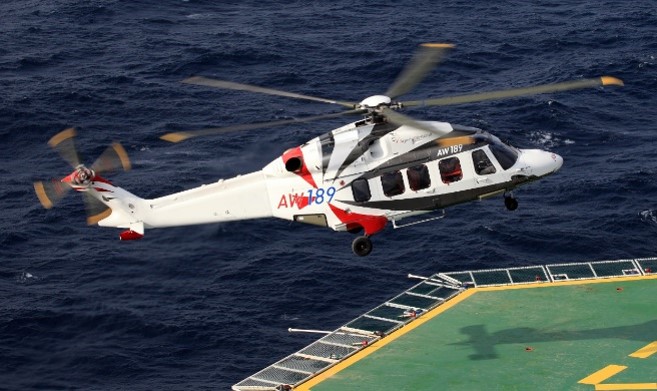}
\caption{Leonardo's AW189 twin-engine helicopter. \textit{Copyright on the images is held by the contributors. Apart from Fair Use, permission must be sought for any other purpose.}}\label{fig1}
\end{figure}

\begin{figure}[h!]
\centering
\includegraphics[width=0.5\textwidth]{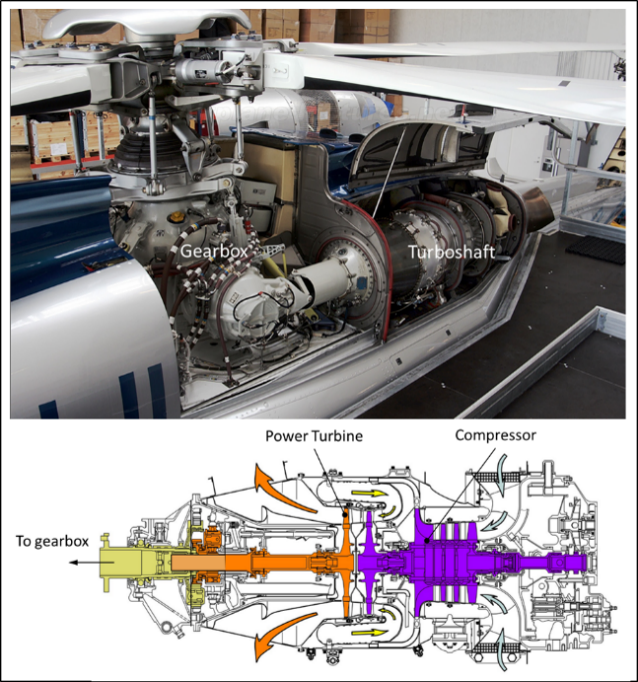}
\caption{Helicopter engine installation. \textit{Copyright on the images is held by the contributors. Apart from Fair Use, permission must be sought for any other purpose.}}\label{fig2}
\end{figure}

Real-time control of helicopter turboshaft engines has always been a complex challenge. Although current numerical techniques enable accurate simulations of engine behavior, their application in real time is impractical due to the significant computational power and time required. A compromise is therefore necessary to develop engine models that can easily be exploited for system control, fault detection and diagnostics studies. For instance, one of the simplest approaches is to obtain a simplified real-time engine simulation by creating piecewise linear state space perturbation models to cover the entire operating range and to support operational needs \cite{duyar1995simplified, litt1989real, merrill1984hytess, duyar1994simplified}. Still, a significant effort is required to develop and tune those models, due to the need for extensive and expensive flight campaigns, which require specific maneuvers and tests. Moreover, the unavoidable changes in engine performance over time and new flight conditions not considered during development may eventually invalidate a properly fine-tuned model, possibly requiring new data and flights. \newline
Over the past decade, turbomachinery and aircraft engine communities have begun to reconsider their design, manufacturing, and operational processes in response to the exponential growth in the use of smart technologies \cite{xie2021digital}. For instance, Artificial Intelligence (AI) and Machine-Learning (ML) algorithms have started to be integrated into several engineering applications, leading to a shift from either empirical or purely theoretical approaches to increasingly accurate solutions capable of capturing and extracting complex nonlinear connections directly from data, albeit at the expense of a physical description and understanding. The modeling of turbomachinery is no exception. \newline
While many efforts have already been made to harness ML algorithms for many different applications, from the design and control of the air and fuel flow \cite{badihi2009artificial, corsini2021cascade, tieghi2021machine}, to deterioration modeling and fault prediction \cite{tayarani2014dynamic, vatani2015health}, this paper focuses on an ML application for helicopter engine modeling, flight mechanics simulation and performance prediction. Despite the significant progress in this field \cite{asgari2020recurrent, de2020hybrid, zheng2018aero, kiakojoori2016dynamic, inproceedings}, the present work has at least two specificities. The first pertains to the unique type of helicopter turboshaft and its associated drivetrain. The second involves the use of large sets of actual flight data from extensive flight test campaigns of a modern rotorcraft, including a rich set of maneuvers and actual engine flight conditions. The dataset was generated by a prototype of Leonardo’s AW189 (\href{https://helicopters.leonardo.com/it/products/aw189-1}{AW189 Product Page}\footnote{https://helicopters.leonardo.com/it/products/aw189-1}
), an 8-ton Maximum TakeOff Weight (MTOW) twin-engine helicopter (Fig.\ref{fig1}).
\newline
To illustrate the complexity of the engine's operation, Fig. \ref{fig2} presents a typical Leonardo helicopter turboshaft drivetrain and a simple sketch of the engine components. The engine first scoops air through the inlet and increases its pressure within the compressor before it enters the combustion chamber, where fuel is sprayed. The resulting air-fuel mixture then expands in the first turbine, which drives the compressor and is mechanically linked to the same shaft. Before exiting through the exhaust nozzle, the hot gases impinge on a second turbine stage, known as the power turbine. This turbine is mechanically decoupled from the first but is connected to the helicopter rotor through a reduction gearbox. As a result, the two turbines can rotate at different speeds, a key feature of helicopter drivetrain design that allows the rotor to maintain a constant RPM independently of the engine regime. The engine power is thus transferred to the helicopter rotor through a transmission gearbox whose reduction ratio is typically in the order of 100 (achieved in successive stages), from around 30,000 rpm in the power turbine to around 300 rpm in the main rotor.
\newline
Due to the intricacies of the drivetrain and the engine's dynamics, accurate helicopter engine simulations play a crucial role in understanding flight mechanics and control. These simulations aim to reproduce the global aeromechanical behavior of the helicopter, providing insights that are essential for effective design and operation. In this context, a realistic engine model is influenced by both pilot control inputs and other critical variables that are of paramount importance for the correct estimation of the helicopter's handling qualities. Specifically, the engine can be simulated as a Multi-Input Single-Output (MISO) model where the output variable is engine torque ($TRQ$) (directly linked to delivered power) which is closely correlated with fuel flow (or Weight of Fuel, abbreviated as $WF$). In fact, $WF$ is an internal variable of the system, not easily monitorable but under the pilot’s control by means of the cyclic, pedals and collective setting ($COL$), regulating the power delivered to the rotor. In addition, the main rotor speed ($NR$) governor ensures that rotor rpm remain stable with minimal fluctuations. The MISO model presented below, takes as input features $COL$, $NR$ and several other air data and engine parameters which affect the delivered $TRQ$. \newline
This paper details the development steps and compares the prediction results of various supervised MISO data-driven models. The models are trained to take as input a specific set of engine and environment variables, known to influence engine behavior but not easily included in physics-based models (either for a specific time-step or for a specific time window). Finally, the model outputs the desired $TRQ$ variable, demonstrating that data-driven engine models can exploit a wider range of parameters than standard transfer function-based approaches. Specifically, we explore two different Neural Networks (NN) architectures: Feed-Forward Neural Network (FFNN) \cite{goodfellow2016deep} and Long-Short-Term Memory (LSTM) recurrent network \cite{hochreiter1997long, staudemeyer2019understanding}. \newline
The rationale of the work lies in the expectation that data-driven models, being capable of describing the dynamic behavior of nonlinear systems \cite{de2020hybrid, zheng2018aero}, can expand the scope of the currently employed transfer function-based approaches by leveraging a broader set of inputs, identifying all those variables that correlate well with the desired output, the engine torque $TRQ$. Therefore, since data-driven engine models can exploit a wide range of parameters describing the turbomachinery, trained schemes can be used to effectively simulate nonlinear effects among different engines, operating regimes, and helicopter models. This may allow for an accurate dynamic simulation over the entire flight envelope of helicopter turboshaft performance, and may also significantly improve their design, optimization, and maintenance processes \cite{kiakojoori2016dynamic}. \newline
By following the same line of reasoning, we also apply Sparse Identification of Nonlinear Dynamics\cite{brunton2016discovering} to derive from the available data a low-dimensional dynamical model describing the relationship between $WF$ and $TRQ$, i.e. a Single-Input Single-Output (SISO) model. The SINDy methodology identifies the smallest number of terms needed to explain the data without a detailed prior physical knowledge, still leading to interpretable governing equations, which may enrich the understanding of this complex system. Unlike other data-driven machine-learning approaches, sparse identification leverages a unique balance in model complexity and descriptive ability, which is essential in the aviation engine field to meet the highest level of model robustness and validation.

\section{Methodology}
This section provides a brief description of the adopted Neural Network architectures for the MISO model, followed by some theoretical background of SINDy. The last subsection provides some details on the Leonardo's AW189 dataset and on the training setup. 

\subsection{Neural Network Architectures}
For the Multi-Input Single-Output models, we implemented two distinct Neural Network architectures using the PyTorch library \cite{paszke2019pytorch} to assess their respective advantages and limitations for modelling the engine dynamics.\newline
The first model is based on a Feed-Forward Neural Network (FFNN)\cite{goodfellow2016deep, lecun2015deep}, which establishes direct correlations between the input features of a specific time step and the output feature at the same time step. FFNNs are structured with one or more hidden layers between the input and output layers, where each hidden layer consists of neurons that are fully connected to the neurons in the previous layer. Information flows through these connections, processed by weights, biases, and nonlinear activation functions. During training, the network parameters are optimized through backpropagation and gradient descent, with the aim of minimizing the loss function. FFNNs are generally efficient in capturing non-linear relationships between input and output variables, but they usually lack the ability to model temporal dependencies, making them less suitable for time series data.\newline
To address this limitation, we also explored a more complex model architecture using Long Short-Term Memory (LSTM) networks. LSTMs are a specialized form of recurrent neural networks (RNNs) and are designed to capture long-term dependencies in sequential data, making them particularly effective for time series. Unlike FFNNs, which process each input independently, LSTMs have memory cells that allow the network to retain information across multiple time steps. The network is fed input time histories (t-hist) in fixed-length sequences, referred to as lookback windows, and predicts an output sequence of the same length. By passing information from one time step to the next, the LSTM can learn temporal patterns and dependencies within the time series. However, this comes at the cost of increased complexity and longer training times compared to more standard FFNN architectures.\newline
In our implementations, we used the Mean Squared Error (MSE) between the predicted and actual output as the loss metric for both the architectures. Details on the input and output features and on the dataset are provided in the following sections.

\subsection{Sparse Identification of Non-Linear Dynamics}
Recent advancements in data-driven parsimonious modeling of system dynamics are paving the way for the development of interpretable ML models. For example, the Sparse Identification of Nonlinear Dynamics (SINDy) approach [18] uses sparse regression to find the smallest number of terms required to model the dynamics of a given system from a library of prescribed candidate functions. Up to now, SINDy algorithm has extensively been adopted in the literature for many different applications \cite{schaeffer2017learning, sorokina2016sparse, dam2017sparse, narasingam2018data, quade2018sparse, loiseau2018sparse, hoffmann2019reactive, lai2019sparse, zhang2018robust}, leading to very promising results in terms of accuracy and interpretability. \newline
In the present work, we employ PySINDy\cite{de2020pysindy, kaptanoglu2021pysindy}, a comprehensive Python package for sparse identification of nonlinear dynamics, to assess the SINDy approach with our AW189 engine database. The goal is to understand the advantages and limitations of this methodology when applied to real data from extensive flight campaigns. However, as a first step, we restrict the analysis to a Single-Input Single-Output (SISO) model, focusing specifically on the relationship between fuel flow ($WF$) and engine torque ($TRQ$), where $WF$ is a control variable that directly influences $TRQ$. This makes the $WF$-$TRQ$ relationship a natural candidate for benchmarking the "SINDy with control" (SINDyc) algorithm \cite{brunton2016sparse}, as this empirical relationship governing $WF$ and $TRQ$ is well documented in the field of engine modeling. The "SINDy with control" algorithm, hereafter referred to simply as SINDy, is briefly described below.\newline
Consider a nonlinear dynamical system like
\begin{equation}
\frac{d\mathbf{x}}{dt} = f(\mathbf{x},\mathbf{u})\label{eq2}\\
\end{equation}
where $\mathbf{x}$ is the state vector collecting all the state variables and $\mathbf{u}$ is the input vector collecting all the control variables. Suppose we have $m$ snapshots of both the state and input vectors arranged in two data matrices $\mathbf{X}$ and $\mathbf{U}$, as
\begin{equation}
\mathbf{X} =\left(\begin{array}{cccc}
                   \vdots & \vdots & \dots & \vdots \\
                   x_{1} &  x_{2} & \dots & x_{m} \\
                   \vdots & \vdots & \dots & \vdots \\
                   \end{array}
             \right)
\end{equation}
\begin{equation}
\mathbf{U} =\left(\begin{array}{cccc}
                   \vdots & \vdots & \dots & \vdots \\
                   u_{1} &  u_{2} & \dots & u_{m} \\
                   \vdots & \vdots & \dots & \vdots \\
                   \end{array}
             \right)
\end{equation}
We may define a library of nonlinear arbitrary candidate functions $\Theta$ of the state and the input including nonlinear cross terms
\begin{equation}
\Theta^T =\left(\begin{array}{ccc}
                   \dots & X & \dots \\
                   \dots & X^{2} & \dots \\
                   \vdots &  \vdots & \vdots \\
                   \dots & U & \dots \\
                   \dots & U^{2} & \dots \\
                   \vdots &  \vdots & \vdots \\
                   \dots & XU & \dots \\
                   \dots & X^{2}U & \dots \\
                   \dots & XU^{2} & \dots \\
                   \vdots &  \vdots & \vdots \\
                   \dots & \sin{X} & \dots \\
                   \dots & \sin{U} & \dots \\
                   \dots & \sin{X}\sin{U} & \dots \\
                   \vdots &  \vdots & \vdots \\
                   \end{array}
             \right)
\end{equation}
and finally solve for the sparse regression coefficients matrix $\Xi$ the equations
\begin{equation}
\frac{d\mathbf{X}}{dt} = \mathbf{\Xi \Theta^T}(\mathbf{X}, \mathbf{U})\\
\end{equation}
by assuming that the input $\mathbf{U}$ corresponds to an external forcing.\\
Note that the sparsity of the coefficients matrix $\mathbf{\Xi}$ is key to obtaining an easily readable and interpretable form of the original system and is assured by the assumption of a threshold parameter $\epsilon$, so that each coefficient lower than $\epsilon$ is automatically set to zero. The threshold should be treated as a hyperparameter of the SINDy algorithm and its correct tuning is essential for each specific case. The time derivative of the state vector appearing at left-hand side of the previous equation, if not available within the dataset, may be evaluated numerically either by numerical differentiation or directly by the PySINDy package, which uses the total variation regularized derivative under the hood. \\
As anticipated, our goal is to describe the relationship between the fuel flow $WF$ and the engine torque $TRQ$. In this case, we may express the state vector as
\begin{equation}
\mathbf{X_i} =\left(\begin{array}{ccccc}
                   TRQ_{i1} & TRQ_{i2} & TRQ_{i3} &\dots & TRQ_{im_i} \\
                   \end{array}
             \right)
\end{equation}
and the input vector as
\begin{equation}
\mathbf{U_i} =\left(\begin{array}{ccccc}
                  WF_{i1} & WF_{i2} & WF_{i3} &\dots & WF_{im_i} \\
                   \end{array}
             \right)
\end{equation}
where the subscript $i$ refers to the specific maneuver of each flight in the dataset.

\subsection{Dataset and training setup}
\begin{figure}[h]
\begin{center}
\includegraphics{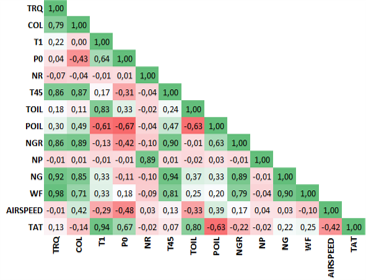}%
\end{center}
\caption{Features correlation matrix}\label{fig3}
\end{figure}

For the development and training of the data-driven MISO and SISO models, we utilized an extensive AW189 engine dataset consisting of time series data for key features of interest, as outlined in Table \ref{tab:table1}.
\begin{table}[h]
  \begin{center}
  \caption{List of variables}
    \label{tab:table1}
    \begin{tabular}{rc}
      \hline
      \textbf{Parameter} & \textbf{Description} \\
      \hline
      $TRQ$ &	Engine Torque [Nm]\\
      $COL$ &	Collective [\%]\\
      $T_{1}$	& Intake Air Temperature [°C]\\
      $T_{45}$	& Gas Turbine Temperature [°C]\\
      $T_{Oil}$	& Oil Temperature [°C]\\
      $P_{Oil}$	& Oil Pressure [psi]\\
      $P_{0}$	& Intake Pressure [psi]\\
      $NR$	& Main Rotor Speed [\%]\\
      $TAT$	& True Air Temperature [°C]\\
      $NP$	& Power Turbine Speed [\%]\\
      $NG$	& Gas Turbine Speed [\%]\\
      $NGR$	& Corrected Gas Turbine Speed [\%]\\
      $WF$	& Fuel Flow [lb/h]\\
      $AIRSPEED$ &	Air Speed [kts]\\
      \hline
    \end{tabular}
  \end{center}
\end{table}
The dataset covers 196 individual flights and spans more than 35 hours of recorded flight time. Each flight test includes a variety of maneuvers, such as cruising, hovering, take-off, bank turns, climbs, descents, reversals, pull-ups, transitions, accelerations, decelerations, autorotations, azimuth forward, sideways, rearward movements, landings, spot turns, sideslips, approaches, quarter maneuvers, taxiing, normal shutdowns and normal Minimum Pitch on Ground (MPOG). However, for the present analysis, some of the listed maneuvers, i.e. autorotations, normal shutdowns, and taxiing, have little or no significance, and were therefore excluded from the dataset. Out of the 196 available flights in the dataset, 180 are used for the models’ development. These are further divided into a training dataset ($D_{train}$) and a validation dataset ($D_{val}$), consisting of 162 and 18 flights, respectively. The remaining 16 flights, which include sweep high-frequency maneuvers (collective and pedal), constitute the test dataset ($D_{test}$) on which all models will be assessed. These flights are never involved in the training phase.\newline
Identical training and cross-validation phases are carried out for the MISO models and for SISO SINDy models by exploiting the Leonardo S.P.A.'s High Performance Computing architecture \textit{davinci-1} (\href{https://www.leonardo.com/en/innovation-technology/davinci-1}{davinci-1 Web Page}\footnote{https://www.leonardo.com/en/innovation-technology/davinci-1}), one of the most powerful supercomputers in the Aerospace, Defence and Security sector.\newline
As ML models rely on statistical analysis and are strongly influenced by the distribution and quality of the data, we performed a correlation analysis to identify highly and poorly correlated variables to be included and/or excluded from the training database (Fig.\ref{fig3}). The correlation between two features ranges between $-1$ and $+1$. Features with a correlation coefficient of $1$ are directly correlated, while those with a correlation coefficient of $-1$ are inversely correlated. Features that are uncorrelated have values close to zero. Since datasets typically contain a large number of features describing each sample, analyzing their correlations may help in selecting the most relevant features and discarding those that are not needed to build an effective predictive model. For example, highly or poorly correlated features can cause multicollinearity in prediction models, a situation in which predictors are linearly dependent, potentially leading to bias in the results. In such cases, highly or poorly correlated features should be removed from the model's input to reduce information redundancy in the data and potentially improving the model performance.\newline
Table \ref{tab:table2} summarizes the final set of variables included into the MISO input-output feature set after the analysis of Fig.\ref{fig3}. It is important to note that some variables, although being good candidates, are not included in the input feature set because they are not normally available. For instance, variables like $T_{45}$, $NG$, $T_{Oil}$ and $P_{Oil}$ are engine internal variables and and are generally inaccessible and unmonitorable under standard conditions.\newline
\begin{table}[h]
  \begin{center}
  \caption{MISO models input-output features set}
    \label{tab:table2}
    \begin{tabular}{cc}
      \hline
      \textbf{Input Features} & \textbf{Target Feature} \\
      \hline
      $COL$, $T_{1}$ ,$P_{0}$, $NR$, $AIRSPEED$ & $TRQ$ \\
      \hline
    \end{tabular}
  \end{center}
\end{table}
The accuracy of all developed models is evaluated using the relative Mean Absolute Error ($rMAE$) index, where a lower value indicates better performance. This score is calculated for each maneuver of interest in each flight using the following formula:
\begin{equation}
rMAE_{i}=\frac{MAE_{i}}{\overline{TRQ_{j}}}\label{eq1}
\end{equation}
where $MAE_{i}$ is the Mean Absolute Error of the $i-th$ maneuver of the $j-th$ flight and $\overline{TRQ_{j}}$ is the average $TRQ$ of that specific flight. Then, to provide an overall index to compare the accuracy of the training phases of each developed model, we first calculate the $rMAE_{j}$, i.e., the average of the $rMAE_{i}$ scores for the $j-th$ flight and next global $rMAE$ score is obtained by averaging all the $rMAE_{j}$. 

\section{Results and Discussion}
To evaluate the capabilities of the developed data-driven models in describing the engine's dynamic behavior, the following sections compare the results obtained during the testing phase on the $D_{test}$ dataset, consisting of 16 flights. The best and worst results are presented by plotting the model predictions against the actual $TRQ$ values for these flights, each of which is uniquely identified by an ID (e.g., ID1, ID2, up to ID16). All the data in the figures are scaled in the range $\left[0, 1\right]$ with a Min-Max scaler.

\subsection{MISO Neural Network Results}
\begin{figure}[h!]
\centering
\includegraphics[width=\textwidth]{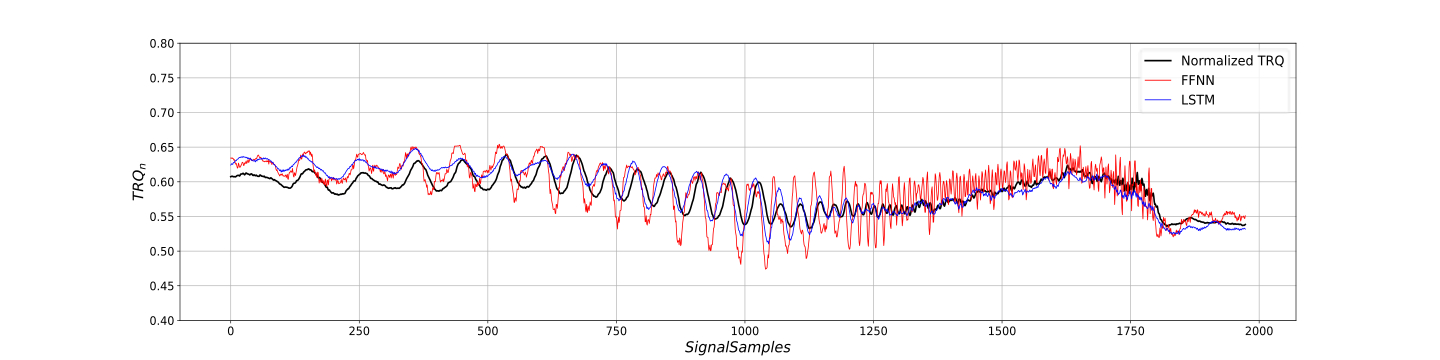}
\caption{Comparison of FFNN and LSTM predictions of normalized $TRQ$ for a maneuver of Flight ID1 in the $D_{test}$ dataset}\label{fig4}
\end{figure}

\begin{figure}[h!]
\centering
\includegraphics[width=\textwidth]{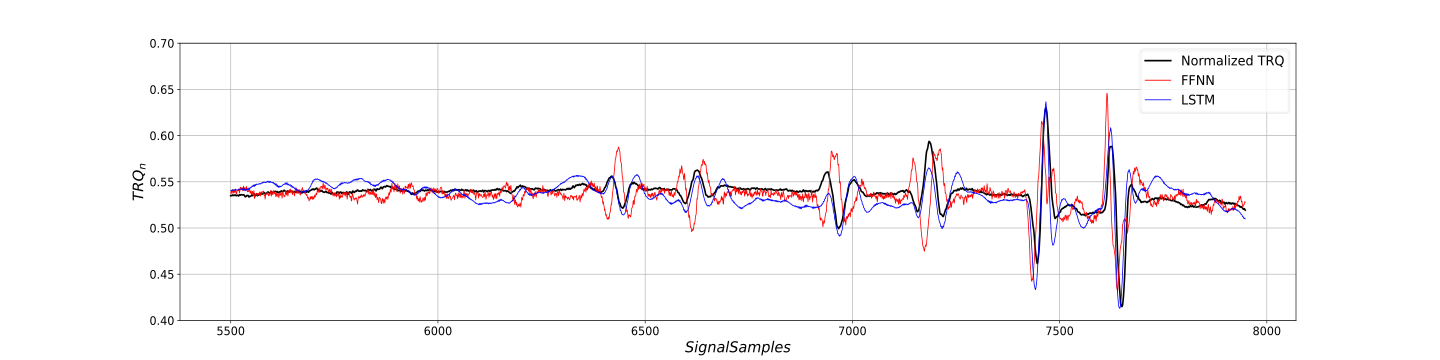}
\caption{Comparison of FFNN and LSTM predictions of normalized $TRQ$ for a maneuver of Flight ID6 in the $D_{test}$ dataset}\label{fig5}
\end{figure}

\begin{figure}[h!]
\centering
\includegraphics[width=\textwidth]{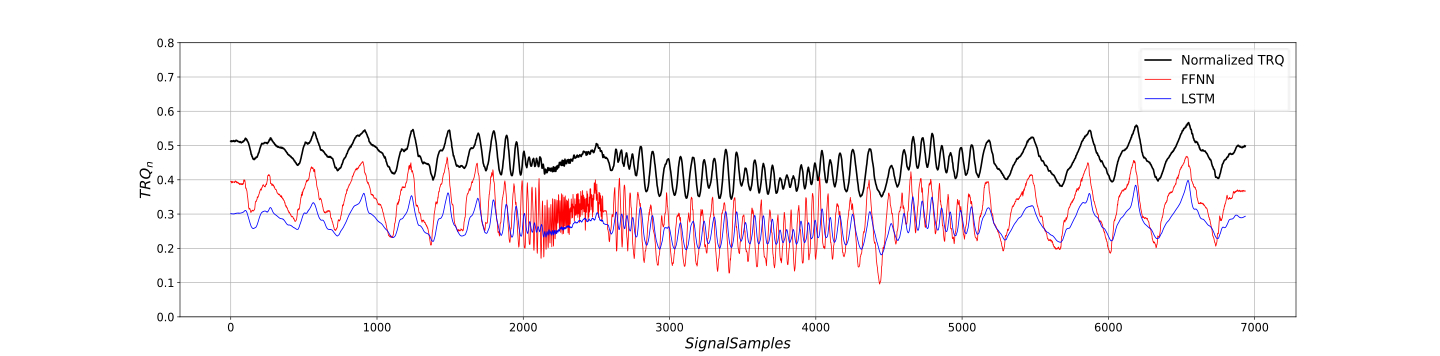}
\caption{Comparison of FFNN and LSTM predictions of normalized $TRQ$ for a maneuver of Flight ID11 in the $D_{test}$ dataset}\label{fig6}
\end{figure}

\begin{figure}[h!]
\centering
\includegraphics[width=\textwidth]{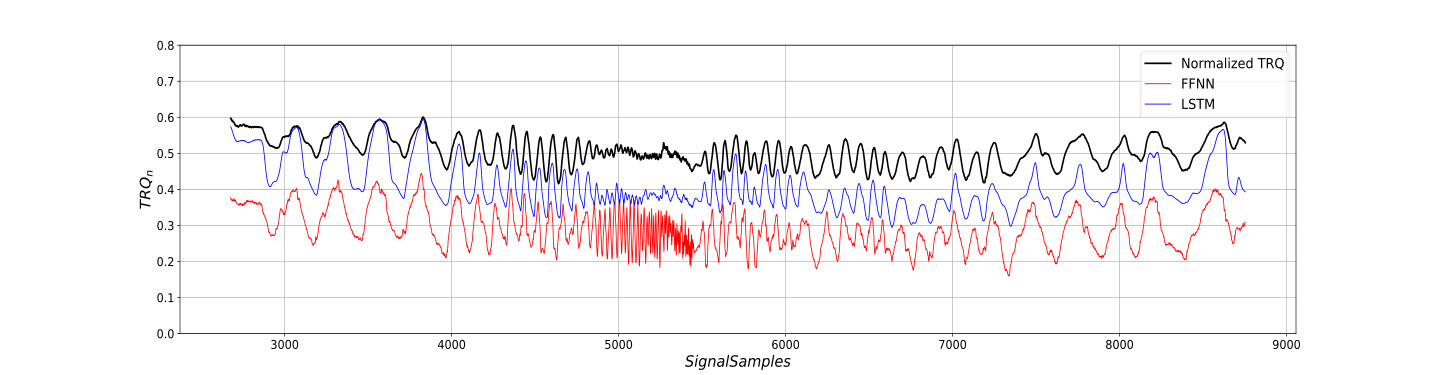}
\caption{Comparison of FFNN and LSTM predictions of normalized $TRQ$ for a maneuver of Flight ID10 in the $D_{test}$ dataset}\label{fig7}
\end{figure}

\begin{figure}[h!]
\centering
\includegraphics[width=\textwidth]{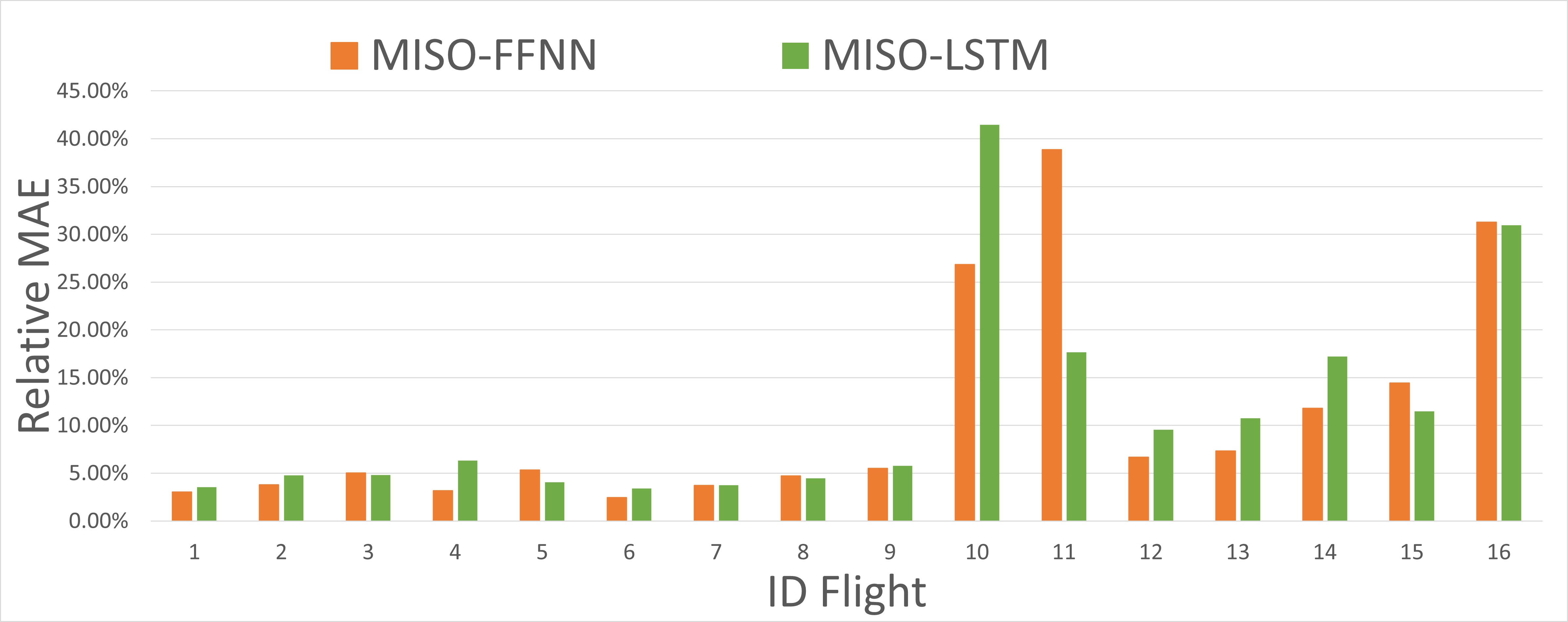}
\caption{MISO models $rMAE$ score per single test flight}\label{fig8}
\end{figure}

\begin{figure}[h!]
\begin{center}
\includegraphics{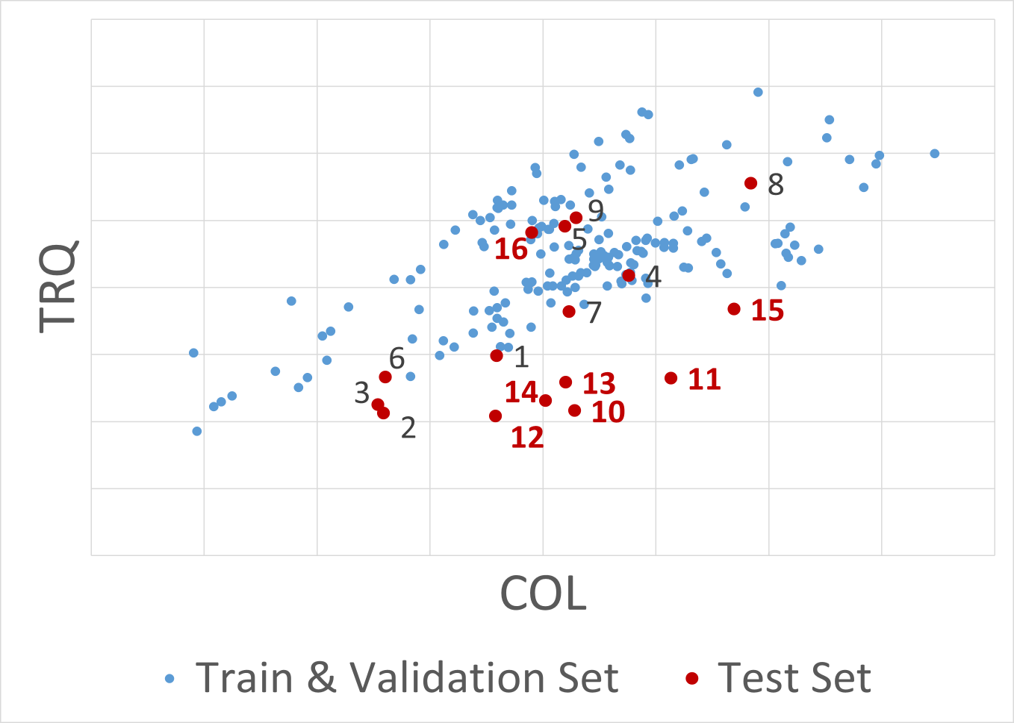}
\end{center}
\caption{Scatter plot of average $COL$ and $TRQ$ values for training/validation and test datasets. Red labels highlight the ID-FLIGHTS in the test set with hight $rMAE$ score.}\label{fig9}
\end{figure}

\begin{figure}[h!]
\begin{center}
\includegraphics[width=5in]{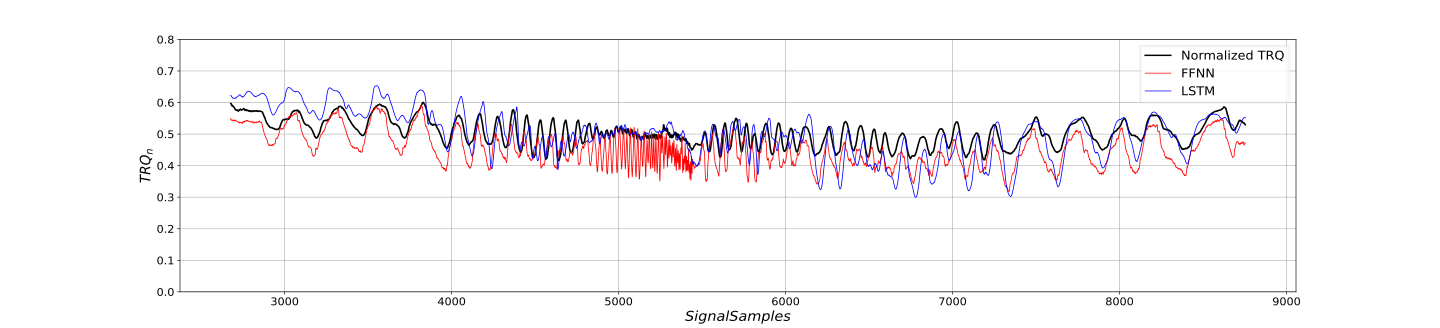}
\end{center}
\caption{Comparison of FFNN and LSTM predictions of normalized $TRQ$ for a maneuver of Flight ID10 in the $D_{test}$ dataset after MISO models re-training.}\label{fig10}
\end{figure}

Table \ref{tab:tab3} details the training setup of the two MISO data-driven models tuned through a Grid Search Algorithm \cite{bergstra2012random} to determine the most suitable network structures, while Table \ref{tab:table4} summarizes the overall $rMAE$ indices for the training phase of each model. The best FFNN model consists of 4 hidden layers of 24 neurons each and the best LSTM model encodes 3 recurrent layers and 6 features in hidden state.\newline
\begin{table}[h]
  \begin{center}
  \caption{MISO models training setup}
    \label{tab:tab3}
    \begin{tabular}{rcc}
      \hline
      \textbf{Hyperparameter} & \textbf{FFNN Model} & \textbf{LSTM Model} \\
      \hline
      Activation Function & ReLU	& ReLU \\
      Optimizer & RMSprop	& ADAM \\
      Loss Function & $MSE$	& $MSE$ \\
      Weights Initialization & Xavier Init & -- \\
      Batch Size & 64 & 64 \\
      N° Epochs & 500 & 100 \\
      Learning Rate & \(1x10^{-4}\) & \(5x10^{-4}\) \\
      Lookback & -- & 20 \\
      \hline
    \end{tabular}
  \end{center}
\end{table}
\begin{table}[h]
  \begin{center}
  \caption{Overall Training mean relative $MAE$ per model}
    \label{tab:table4}
    \begin{tabular}{rc}
      \hline
      \textbf{Model} & \textbf{Overall Mean $rMAE$} \\
      \hline
      MISO FFNN & 0.0368 \\
      MISO LSTM & 0.0317 \\
      \hline
    \end{tabular}
  \end{center}
\end{table}
Fig.\ref{fig4} and Fig.\ref{fig5} present a pair of the best results obtained for two different maneuvers of two different flights from the test set. In particular, Fig.\ref{fig4} refers to flight ID1 and compares the $TRQ$ predictions of FFNN ($rMAE_{FFNN}=3.75\%$) and LSTM ($rMAE_{LSTM} = 2.34\%$) with the normalized actual $TRQ$. Although in terms of $rMAE$ per maneuver the results obtained by the two MISO models may seem comparable, it is evident from the graph that FFNN is unable to capture the engine dynamics. In fact, its weakness emerges especially for maneuvers (such as the collective sweep) in which the dynamic response of the motor is intentionally excited. A second problem is the lack of smoothness of the reconstructed $TRQ$ time series, which has a general "noisy" and "spiky" behavior. The lack of smoothness is a direct consequence of network structure and architecture, where each time instant is treated in isolation without taking into account the time history of the input variables. On the contrary, the predictions of LSTMs fit the original signal better; their feedback connections allow them to handle the sequential nature of time series data, capturing long-term patterns and dependencies, which makes them ideal for predicting time series data.
Fig.\ref{fig5} is a second example of a comparison between the prediction results of MISO models and the normalized real $TRQ$. Again, it is evident that the LSTM network ($rMAE_{LSTM}=1.82\%$) provides a better prediction of the dynamic behavior of the $TRQ$ with respect to the FFNN ($rMAE_{FFNN} =2.21\%$), which continues to produce a very noisy result and, more importantly, fails to capture sudden spikes in the measured TRQ.\\
Fig.\ref{fig6} and Fig.\ref{fig7} show a pair of the worst results for two different maneuvers in the test set The dynamic component appears to be well predicted, at least by the LSTM model, but the predicted values seem to be consistently shifted by a constant. This suggests that the static component of the dynamics is not accurately captured.
From Fig.\ref{fig8}, which shows the $rMAE$ scores per single test flight, it can be easily seen that, while for flights between ID1 and ID9 the $rMAE_{j}$ is acceptable and comparable to the training and testing scores, for the remaining flights (ID10 to ID15) the $rMAE_{j}$ is significantly larger, with $rMAE_{j}$ scores as high as 42.42\%, all associated to a static offset of the predicted $TRQ$.\\
The causes of this behavior can be related to errors and uncertainties in the collected flight data. The data from flights ID1 to ID9 pertains to flight tests conducted in 2016, while flights ID10 to ID15 correspond to tests carried out in 2018. Therefore, we may not exclude that the instrumentation may have been replaced or calibrated in a different way, or that something changed in the engine behavior. Another cause may be the lack of essential variables not included in the input feature set. One such feature could be the helicopter climb rate, which, for the same $COL$, leads to a different $TRQ$ response. This hypothesis is supported by the relationship between the average $COL$ and average $TRQ$ per flight (Fig.\ref{fig9}): a discrepancy may be observed between the distribution of the training dataset and the one of the test dataset. In fact, the majority of flights with the highest relative $MAEs$ do not fall within the distribution of training flights. Hence, neural networks struggle to accurately predict $TRQ$ because the training data do not cover these behaviors.
To achieve greater accuracy in $TRQ$ prediction on $D_{test}$ flights, a new training phase of MISO models is performed by including 11 $D_{test}$ flights in $D_{train}$. The 11 test flights added to $D_{train}$ include both flights with ID1 to ID9 (for which the MISO models already work quite well) and flights with ID10 to ID15, which are not part of the initial training set.\\
By incorporating the test flights into the training set, MISO model performance improves; in fact, the scores obtained in the test phase after models re-training phase, go from a $rMAE_{FFNN}=13.50\%$ to a $rMAE_{FFNN}=5.78\%$ and from a $rMAE_{LSTM}=8.32\%$ to a $rMAE_{LSTM}=4.06\%$.\\
The performance improvements of the two MISO models are even more evident when looking at Fig.\ref{fig10}, which shows the same ID10 flight maneuver as in Fig.\ref{fig7}. Even though the predictions are now mostly superimposed to the normalized real $TRQ$, the FFNN model still fails in properly predicting high-frequency and small amplitude $TRQ$ oscillations for sweep flights.


\subsection{SINDy Results}
As a first result of the SINDy model training over the $D_{train}$ dataset, by selecting a polynomial candidate functions library, we obtained the following relation:
\begin{equation}
\dot{TRQ}(t) = -a - b \, TRQ(t) + c \, WF(t)
\end{equation}
where \emph{a}, \emph{b} and  \emph{c} are all positive coefficients and t is time. In fact, this result is not unexpected. In standard transfer-function modeling approaches, the  $WF$-$TRQ$ relation is modeled by a transfer function as follows,
\begin{equation}
\frac{TRQ(s)}{WF(s)}=\frac{\mu}{1+\tau s}
\end{equation}
which may be easily reconducted to a first order ODE in time-domain. Even though this result shows how SINDy is able to correctly recover the system dynamics from data without previous knowledge of the underlying physics, the first order equation (9) is not so useful in terms of accuracy, as intuitively shown by the overall mean $rMAE$, reported in Table 5. The simulations are obtained for each maneuver in $D_{test}$ by using as initial conditions the value of $TRQ$ at the beginning of the maneuver itself, and then integrating in time via the time integration algorithm provided by PySINDy package. 
A step further is obtained by forcing a second order model for SINDy prediction. The strategy is to modify the state and input vectors by including the time derivatives of $TRQ$ and $WF$:
\begin{equation}
\mathbf{X_i} =\left(\begin{array}{ccccc}
                   TRQ_{i1} & TRQ_{i2} & TRQ_{i3} &\dots & TRQ_{im_i} \\
                   \dot{TRQ_{i1}} & \dot{TRQ_{i2}} & \dot{TRQ_{i3}} &\dots & \dot{TRQ}_{im_i} \\
                   \end{array}
             \right)
\end{equation}
\begin{equation}
\mathbf{U_i} =\left(\begin{array}{ccccc}
                   WF_{i1} & WF_{i2} & WF_{i3} &\dots & WF_{im_i} \\
                   \dot{WF_{i1}} & \dot{WF_{i2}} & \dot{WF_{i3}} &\dots & \dot{WF}_{im_i} \\
                   \end{array}
             \right)
\end{equation}
so as to look for a system of equations instead.\\
In this case, the time-derivatives of the state and input signals are evaluated by numerical differentiation and are used to train SINDy model, leading to the following result
\begin{equation}\label{sindy2nd}
\left\{
\begin{array}{l}
\dot{TRQ}(t) = a^\prime TRQ(t) \\
\ddot{TRQ}(t) = b^\prime \dot{TRQ}(t) + c^\prime \dot{WF}(t)
\end{array}
\right.
\end{equation}
where the first is clearly a dummy equation $a\prime=1$ which follows from the SINDy algorithm and $b^\prime$ and $c^\prime$ are positive coefficients. By using the second order SINDy model given by system \ref{sindy2nd} to simulate the $TRQ$ time-histories of the test dataset, a significant improvement is obtained. As reported in Table \ref{tab:table5}, we obtain a reduction of one order of magnitude with respect to the first order model in terms of overall mean relative $MAE$. In this case, an additional initial condition is required on the $\dot{TRQ}$, while the new input variable is the time-derivative of $WF$.  
\begin{table}[h]
  \begin{center}
  \caption{Overall Training mean relative $MAE$ per SINDy model}
    \label{tab:table5}
    \begin{tabular}{rc}
      \hline
      \textbf{Model} & \textbf{Overall Mean $rMAE$} \\
      \hline
      SISO SINDy $1^{st}$ Order & 0,137 \\
      SISO SINDy $2^{nd}$ Order & 0,016 \\
      \hline
    \end{tabular}
  \end{center}
\end{table}
\\
At first glance, the very large difference between the two SINDy models may appear counterintuitive since the second equation of system \ref{sindy2nd} is easily traced back to a first order model by integrating in time. Actually, we obtain
\begin{equation}
\int{\ddot{TRQ}(t)}=\int{-b^\prime\dot{TRQ}(t)dt}+\int{c^\prime\dot{WF}(t)dt}
\end{equation}
leading to 
\begin{equation}\label{sindy2to1}
\dot{TRQ}(t)=-b^\prime TRQ(t)+c^\prime WF(t)+const
\end{equation}
which is the same expression as equation (9). However, the coefficients resulting from the two different training processes are quite different and the results provided by the second order SINDy model are sufficiently accurate even for $D_{val}$.\newline
Before proceeding with the discussion of the result for the validation dataset, it should be noticed that the SINDy model differs for a very important aspect from the FFNN and LSTM models presented in the previous paragraphs. The SINDy model takes only an internal variable of the system as input, while ignoring the real pilot control variables, namely collective ($COL$) and main rotor speed ($NR$). It follows that a direct comparison of SINDy $2^{nd}$ order scores with those of FFNN and LSTM may be misleading, as equation \ref{sindy2to1} pertains only to a single engine component, providing a direct relation between $WF$ and $TRQ$. Still, it is interesting to discuss the following results to demonstrate SINDy's potential to extrapolate accurate physical models from normal flight data. \\
\begin{figure}[h!]
\begin{center}
\includegraphics[width=5in]{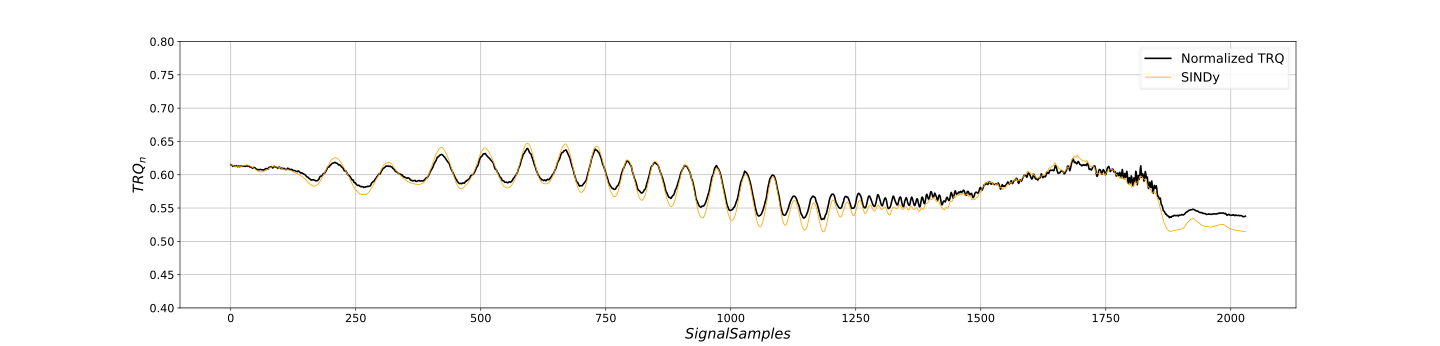}
\end{center}
\caption{SINDy $2^{nd}$ order prediction of normalized $TRQ$ for a maneuver of Flight ID1 in the $D_{test}$ dataset.}\label{fig11}
\end{figure}
\begin{figure}[h!]
\begin{center}
\includegraphics[width=5in]{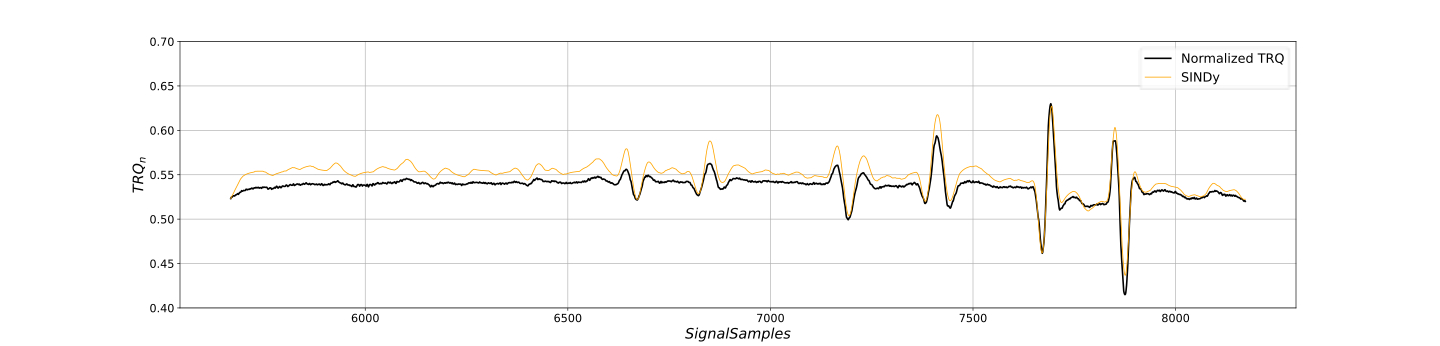}
\end{center}
\caption{SINDy $2^{nd}$ order prediction of normalized $TRQ$ for a maneuver of Flight ID6 in the $D_{test}$ dataset.}\label{fig12}
\end{figure}
Fig.\ref{fig11} to Fig.\ref{fig14} present the same flight maneuvers discussed in subsection 3.1 for the MISO neural networks. It is clear that the simulation results obtained from equation (13) are highly accurate, reproducing very well TRQ dynamics with only slight under/overestimations of some oscillation peaks. Again, an estimate of the total error is given by the rMAE score per test flight, which in this case is always lower than $3\%$. As anticipated above, these results demonstrate the potential of the SINDy approach in extrapolating a simple and interpretable dynamic model even without a deep understanding of the underlying physics of the system. However, in this case, the transition to the second order is crucial to achieve more than satisfactory accuracy. This behavior, in our experience, is not easily predictable, as $WF$-$TRQ$ relationship is typically modeled using transfer functions associated to $1^{st}$ order ODE (see equation 10). It is likely that the information carried by the temporal derivative of fuel flow, the new control variable for system of equations (13), together with the additional initial condition on $\dot{TRQ}$, may constitute the key for the improvement achieved in terms of model performance.
\begin{figure}[h!]
\begin{center}
\includegraphics[width=5in]{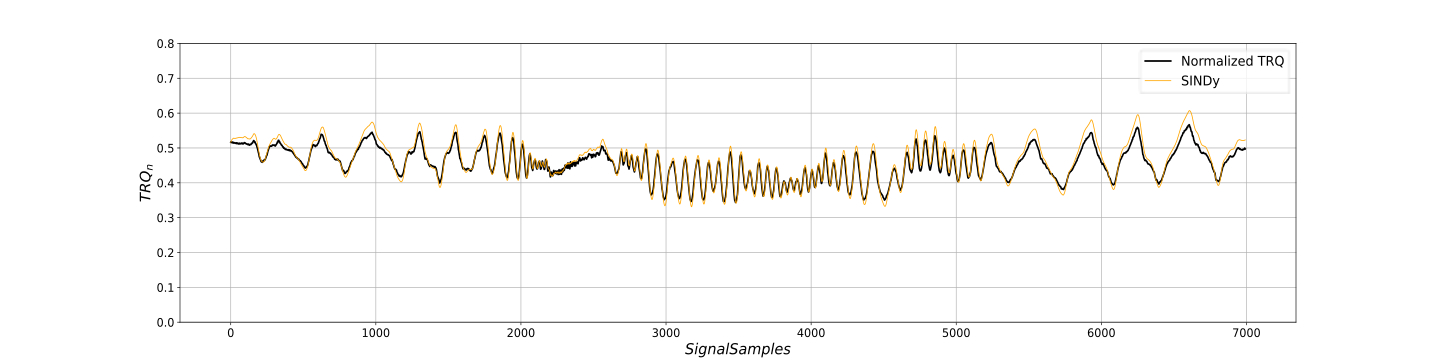}
\end{center}
\caption{SINDy $2^{nd}$ order prediction of normalized $TRQ$ for a maneuver of Flight ID11 in the $D_{test}$ dataset.}\label{fig13}
\end{figure}
\begin{figure}[h!]
\begin{center}
\includegraphics[width=5in]{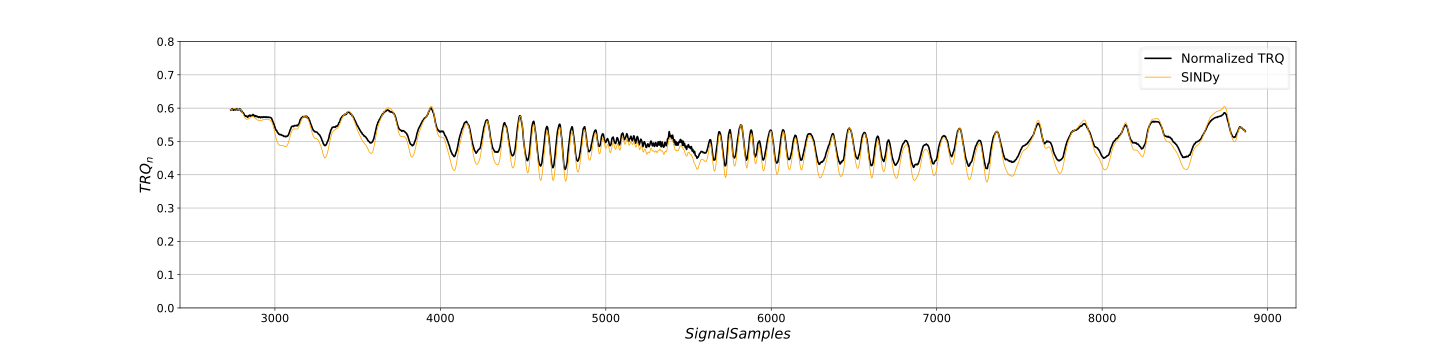}
\end{center}
\caption{SINDy $2^{nd}$ order prediction of normalized $TRQ$ for a maneuver of Flight ID10 in the $D_{test}$ dataset.}\label{fig14}
\end{figure}

\section{Conclusions}
In this paper, three different supervised data-driven approaches have been presented with the goal of developing a dynamic turbo-shaft engine model to extend the scope of commonly used transfer function-based models. The developed data-driven engine models, based on a database of real flight tests of Leonardo's AW189 prototypes, allow for accurate predictions of $TRQ$ over the entire flight envelope of the helicopter, demonstrating that this methodology may be effectively used to simulate nonlinear behaviors of different engines.\\
Specifically, two different MISO neural network architectures are developed, namely a FFNN and a LSTM network, which take as input the time-histories of engine and environmental variables to predict the desired engine torque. The results show that the FFNN is not able to ensure very accurate predictions, which are also characterized by strong oscillations, not acceptable for engine control. On the other hand, the LSTM, being a recurrent network, yields to better predictions of $TRQ$ dynamic behavior since its output is based on a time window that improves performance in terms of signal smoothness. \\
In addition, with the intent of developing an interpretable ML dynamical model, we use SINDy approach to derive a very accurate model from the available flight data even without a deep understanding of the underlying physics of the system. The obtained results show that the SISO SINDy model can effectively reproduce the $TRQ$ dynamics once a second order model is forced in SINDy algorithm. As a future step of this work, considering the encouraging results obtained with SINDy, we envisage extending this methodology to obtain a complete engine model for predicting torque from the real pilot control inputs.

\newpage

\bibliographystyle{unsrt}
\bibliography{paniccia_et_al_25}
\end{document}